\pdfoutput=1

\documentclass[11pt]{article}

\usepackage[a4paper,margin=1in]{geometry}
\usepackage{amsmath,amssymb}
\usepackage{graphicx}
\usepackage{placeins}
\usepackage{xcolor}
\usepackage[hidelinks]{hyperref}
\usepackage{array}
\usepackage[authoryear,round]{natbib}
\usepackage{authblk}
\setcitestyle{authoryear,round,semicolon}
\newcolumntype{P}[1]{>{\raggedright\arraybackslash}p{#1}}
\newcommand{\keywords}[1]{\par\noindent\textbf{Keywords:} #1\par\medskip}
\title{Bayesian Expected Uncertainty Reduction (B-EUR) Model:\\
A Computational Account of What Makes Design Options Worth Trying}

\author[1]{Shimon Honda}
\author[1]{Takuma Miyaguchi}
\author[1]{Koji Koizumi}
\author[1]{Takanori Sano}
\author[2]{Tristan Briard}
\author[1]{Hideyoshi Yanagisawa\thanks{Corresponding author: \href{mailto:hide@mech.t.u-tokyo.ac.jp}{hide@mech.t.u-tokyo.ac.jp}}}

\affil[1]{Department of Mechanical Engineering, Graduate School of Engineering, The University of Tokyo, 7-3-1 Hongo, Bunkyo-ku, Tokyo 113-8656, Japan}
\affil[2]{Laboratoire Conception de Produits et Innovation (LCPI), Arts et M\'etiers ParisTech, 151 Boulevard de l'H\^opital, 75013 Paris, France}
\date{}

\begin{document}
\maketitle

\begin{abstract}
This paper proposes the Bayesian Expected Uncertainty Reduction (B-EUR) model, which formalizes the value of trying a candidate design action as its expected reduction of epistemic uncertainty about action--outcome relations. The model addresses one part of the Uncertainty Driven Action (UDA) model's open question concerning how changes in uncertainty perception determine action selection. We examine two environmental properties: generalizability, or how far knowledge from one trial extends to neighboring candidates, and outcome discriminability, or how clearly differences among outcomes can be distinguished.

We tested the model through simulations and human experiments using a graph-shape guessing task that isolates learning about action--outcome relations under a limited trial budget. Epistemic value followed an inverted-U-shaped relationship with generalizability and increased with outcome discriminability in the simulations. In the human experiments, the subjective value of trying and enjoyment followed inverted-U-shaped relationships with generalizability, while choice behavior reflected both properties.

The B-EUR model provides a computational account of candidate-action evaluation within uncertainty-driven design activity and offers implications for constructing prototype sets, framing design problems, and organizing feedback to support informative exploration.
\end{abstract}


\section{Introduction}

\subsection{Trying design options under uncertainty}

Design processes involve uncertainty. In the early stages of product development, designers often need to make decisions with limited information \citep{Gembarski_2021,Gray_2022}. Under such uncertainty, designers need to try design options and learn from their outcomes. Trial and error helps designers expand their knowledge of the design space and infer the properties of untested candidates \citep{Erat_2008}. Experimentation and iterative debugging also relate to organizational learning and exploration driven by unexpected failures \citep{Koning_2022,Alblas_2021}. More broadly, design exploration has been described as a process of generating and evaluating previously unconsidered alternatives beyond a predefined solution space, while iteratively clarifying incomplete or evolving requirements \citep{Navinchandra1991,Logan_Smithers_1993}. In this study, we focus on one aspect of this broader process: learning action--outcome relations through trials.

At the same time, not all trials are equally useful. Whether a trial leads to useful learning depends on the proximity among design options and on whether design attributes involve nontrivial interactions \citep{Erat_2008}. The number of trial-and-error iterations also does not necessarily improve learning or performance \citep{Abhyankar_2014}. To understand and support design exploration, we need to clarify which environmental properties increase the value of trying a design option.

Prior design research has suggested that uncertainty plays a central role in driving design exploration. In engineering design, uncertainty is often divided into epistemic uncertainty, which arises from a lack of knowledge and can be reduced through information, and aleatory uncertainty, which arises from inherent randomness and cannot be reduced in principle \citep{Reneke_2010,Dai_2003}. The Uncertainty Driven Action (UDA) model conceptualizes design activity as a progression through information, knowledge-sharing, and representation actions linked by uncertainty perception \citep{Cash_2017,Cash_2018}. The model proposes that changes in the level and nature of uncertainty perception determine action selection and progression. However, it remains unclear exactly how these changes determine action selection. Design research has also treated uncertainty not only as something to reduce but also as a resource for opening inquiry \citep{Cash_2017,Dyer_2021,Giaccardi_2024,Stompff_2022,Epp_2024}.

Accordingly, an unresolved issue is how changes in uncertainty perception lead to the selection of candidate design actions. To address one part of this issue, we focus on actions that produce observations about action--outcome relations and use a computational approach to examine how environmental properties shape the value of trying these actions.

\subsection{Generalizability and outcome discriminability}

We define two environmental properties that shape the value of trying a design option: generalizability and outcome discriminability. First, generalizability refers to the extent to which knowledge gained from one trial extends to neighboring candidates. For example, when feedback from one prototype helps predict responses to similar prototypes, the trial provides information about untested candidates as well \citep{Erat_2008}. If the acquired knowledge does not generalize to other candidates, each trial remains isolated and contributes little to understanding the broader design space. Conversely, if generalizability is too high, different candidates provide highly overlapping information, so the total information gained from exploring multiple candidates becomes limited.

Second, outcome discriminability refers to the extent to which outcome differences among candidates can be clearly distinguished. If multiple design options yield ambiguous or indistinguishable responses, designers may struggle to learn which differences led to which outcomes. In contrast, when outcome differences are sufficiently distinguishable, designers can use trial outcomes to learn action--outcome relations. For example, prototypes that differ only in a barely noticeable trim detail may produce almost indistinguishable user responses, whereas prototypes that differ in a salient element, such as control layout or material texture, may produce clearer differences in usability, reassurance, or perceived quality.

\subsection{Expected uncertainty reduction as epistemic value}

We conceptualize the value of trying a design option as the extent to which the trial is expected to reduce uncertainty about action--outcome relations. Accordingly, we take the position that the value of an action depends on how much uncertainty that particular action is expected to reduce.

This idea connects to active inference \citep{Friston_2015,Friston_2017}, which extends the free-energy principle \citep{Friston_2010} as a theory of brain function to action selection. In active inference, agents select actions based on both pragmatic value, associated with desired outcomes, and epistemic value, associated with expected uncertainty reduction through observation \citep{Parr_2017,Schwartenbeck_2019,Yanagisawa_2025}. Previous theoretical and empirical studies have associated reductions in free energy following recognition or belief updating with positive emotional valence, including interest, and have modeled the corresponding information gain as the valence of epistemic emotions \citep{Yanagisawa_2023,Ueda_2025,Yanagisawa_2025}. Related studies in engineering design have applied information-theoretic formulations to aesthetic shape generation, novelty quantification, and interest and sustained engagement in motion design \citep{Honda_2022,Sasaki_2024,Honda_2025}. Applied to design exploration, a design option has high epistemic value when trying it is expected to update beliefs about action--outcome relations.

We propose the Bayesian Expected Uncertainty Reduction (B-EUR) model, which formalizes the value of a candidate action as expected uncertainty reduction. In relation to the UDA model, B-EUR does not model the full progression and combination of information, knowledge-sharing, and representation actions. Rather, B-EUR specifies a computational mechanism for evaluating candidate actions prior to selection. This allows us to reframe the question of what makes a design option worth trying as the question of which environmental properties increase expected uncertainty reduction.

\subsection{The present study}

This study aims to clarify which environmental properties increase the value of trying a design option using the B-EUR model and human experiments. We focus on generalizability and outcome discriminability, and examine how they affect expected uncertainty reduction, perceived uncertainty, enjoyment of the exploration process, and choices among exploration conditions.

To do so, we use a graph-shape guessing task in which participants infer an unknown one-dimensional function under a limited number of trials. This abstract task excludes pragmatic value, such as obtaining high output values, and isolates design exploration based on epistemic value, namely learning action--outcome relations.

In the graph-shape guessing task, we manipulate the environmental properties as parameters of a Gaussian process \citep{Angel_Deborah_2021,Han_2012}. We map generalizability onto correlation length $l$, which controls how slowly correlations between input values decay with distance. A larger $l$ therefore makes observations generalize more broadly to neighboring inputs. We map outcome discriminability onto amplitude $\sigma$, which controls the prior variance of output values. A larger $\sigma$ increases vertical variation in the graph and makes outcome differences easier to distinguish.

The main contribution of this study is to clarify, through both a computational model and human experiments, which environmental properties increase the value of trying a design option. In doing so, this study addresses one part of the UDA model's open question concerning how changes in uncertainty perception determine action selection. Specifically, this study provides a computational account of how candidate actions can be evaluated in terms of expected uncertainty reduction. To this end, we construct the B-EUR model and use simulations to show how generalizability and outcome discriminability affect epistemic value. We then use the graph-shape guessing task to test how these environmental properties appear in subjective ratings and choices. Finally, we discuss how these findings can inform prototype sets, problem framing, and feedback practices that help designers create exploration conditions for obtaining more informative outcomes from trials.

\section{Bayesian Modeling of Design Exploration}

\subsection{Method}

In this section, we introduce the Graph-Shape Guessing Game as an abstract task for design exploration and describe the B-EUR model, which formalizes expected uncertainty reduction as epistemic value.

\subsubsection{Graph-Shape Guessing Game}

\begin{figure}[!htbp]
\centering
\includegraphics[width=0.85\linewidth]{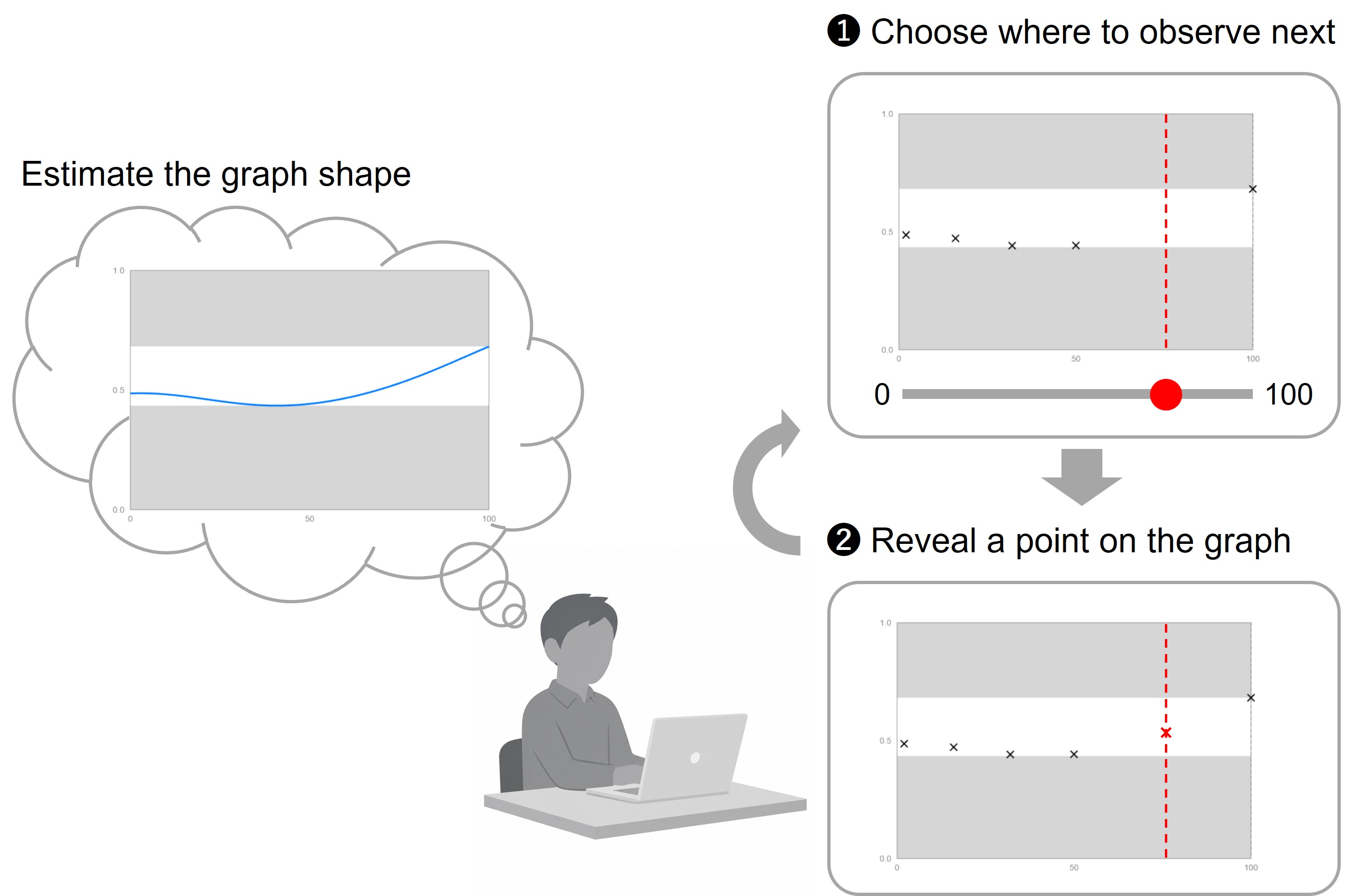}
\caption{
Overview of the Graph-Shape Guessing Game.
Participants estimated the overall graph shape from the observed points, selected the next input location using a slider, and then revealed the output value at that location as a new observed point.
By repeating this process, participants learned the unknown graph shape and reduced uncertainty about the input--output relationship.
The central human illustration in this figure was generated using ChatGPT (OpenAI).
}
\label{fig:graph_shape_game}
\end{figure}

We introduced the Graph-Shape Guessing Game as an abstract model of one aspect of design exploration. In this task, participants and computational agents inferred the overall shape of an unknown one-dimensional function by selecting an input value (x-coordinate) and observing the corresponding output value (y-coordinate) one point at a time. Each trial allowed 10 observations. Figure \ref{fig:graph_shape_game} shows an overview of the Graph-Shape Guessing Game.

This task can be regarded as a variant of spatially correlated Multi-Armed Bandit (MAB) tasks \citep{Wu_2018,Schulz_2020,Wu_2020,Schulz_2018}. In contrast to standard MAB tasks, our task introduced neither preferences for observations, such as obtaining high output values, nor external rewards. The task goal was not to maximize payoff but to learn the relationship between inputs and outputs. Thus, we designed the task to exclude the influence of pragmatic value and isolate exploration based on epistemic value.

This task abstracts one aspect of design exploration: learning how design actions, such as prototyping, evaluation, and review, lead to outcomes such as performance results or user responses. Table \ref{tab:graph_design_mapping} shows the correspondence between elements of the Graph-Shape Guessing Game and design exploration. We needed this abstraction to examine the basic environmental properties that increase epistemic value. If we had used an actual prototyping task, many confounding factors would have emerged, including the context dependence of evaluation due to product category and usage situation \citep{Hu_2010,Berni_2024,De_Maeyer_2011}, pragmatic value such as obtaining high ratings or usefulness \citep{Simon_1988,Ranjan_2018}, and differences in designers' expertise and prior knowledge \citep{Tseng_2011}. To experimentally examine how generalizability $l$ and outcome discriminability $\sigma$ affect epistemic value and perceived uncertainty, we needed an abstract task that could control these confounding factors.

We generated the graphs using Gaussian processes with an RBF kernel as the covariance function \citep{Angel_Deborah_2021,Han_2012}. The covariance between two input values $u$ and $u'$ is given by

\begin{equation}
k(u,u')=\sigma^2 \exp\!\left(-\frac{(u-u')^2}{2l^2}\right).
\end{equation}

Here, $\sigma$ denotes amplitude and $l$ denotes correlation length. We treat $\sigma$ as an environmental parameter that determines outcome discriminability and $l$ as an environmental parameter that determines generalizability. A larger $\sigma$ expands the range of output values and increases differences among observed outputs. A smaller $l$ makes the graph vary more locally, so observations generalize less to neighboring inputs. Conversely, a larger $l$ makes the graph smoother, so observations at one location generalize more broadly to neighboring inputs.

We used this task as a human experimental task and also built a computational agent that performed the same task based on active inference.

\begin{table}[!htbp]
\centering
\caption{Correspondence between the Graph-Shape Guessing Game and design exploration}
\label{tab:graph_design_mapping}
\setlength{\tabcolsep}{4pt}
\begin{tabular}{P{0.28\linewidth}P{0.62\linewidth}}
\hline
Element in the Graph-Shape Guessing Game & Correspondence in design practice \\
\hline
Action of selecting an input value $u$
& Performing a design action, such as selecting a design option to evaluate, prototype, or examine \\

Observed value $f(u)$
& Outcomes or responses obtained from a design action, such as prototype performance, customer feedback, sensory evaluation scores, or review comments \\

Overall graph shape
& The relationship between design options and outcomes, such as the unknown relation between exterior design features and customer responses or evaluations \\

Limited number of observations (10 observations)
& Resource constraints, such as time and budget for prototyping, evaluation, user testing, or review \\

Correlation length $l$ (generalizability)
& The extent to which knowledge gained from one design option generalizes to the outcomes or responses of similar design options \\

Amplitude $\sigma$ (outcome discriminability)
& The extent to which differences in outcomes or responses among multiple design options can be distinguished \\

Epistemic value
& The expected increase in understanding of the relationship between design options and outcomes, including untested options, through prototyping and evaluation \\
\hline
\end{tabular}
\end{table}

\subsubsection{B-EUR Model: Action Selection Based on Expected Uncertainty Reduction}

We propose the Bayesian Expected Uncertainty Reduction model (B-EUR model), which formalizes the value of trying a design action as expected uncertainty reduction. To clarify the relation between design terminology and the active-inference terminology used in the model, we use the term environment to refer to the structure of the design situation in which actions produce outcomes. The UDA model conceptualizes design activity as involving information, knowledge-sharing, and representation actions \citep{Cash_2017,Cash_2018}. The B-EUR model does not distinguish among or model transitions across these action categories. Instead, it abstracts a common action--outcome structure in which a candidate action produces an observation that updates beliefs about action--outcome relations.

In the Graph-Shape Guessing Game, this design situation is abstracted as selecting an input value and observing the corresponding output value. A designer or agent selects an input value, that is, an action $u_t$, at time $t$ and observes an output value $o_{t+1}$. Based on the observation $o_{t+1}$, the agent infers a discretized output state $s_{t+1}$. The learning target is the mapping between the action $u_t$ and the next state $s_{t+1}$, which we denote as

\begin{equation}
B_t = P(s_{t+1}\mid u_t).
\end{equation}

The epistemic value $V_{\mathrm{epi}}(u_t)$ of trying an input value $u_t$ depends on how much uncertainty about the action--outcome mapping $B_t$ the agent expects the observation to reduce. We use $q(\cdot)$ to denote the agent's belief distribution. Specifically, $q(B_t)$ denotes the current belief distribution over the mapping, $q(s_{t+1}\mid u_t)$ denotes the predicted distribution of the next state after choosing action $u_t$, and $q(B_t\mid s_{t+1},u_t)$ denotes the posterior belief distribution over the mapping after observing state $s_{t+1}$. We express this value as the following mutual information $I(B_t;s_{t+1}\mid u_t)$:

\begin{equation}
V_{\mathrm{epi}}(u_t)
=
I(B_t;s_{t+1}\mid u_t)
=
\underbrace{
H[q(B_t)]
}_{\substack{\text{epistemic uncertainty}\\ \text{before observation}}}
-
\underbrace{
\mathbb{E}_{q(s_{t+1}\mid u_t)}
\left[
H[q(B_t\mid s_{t+1},u_t)]
\right]
}_{\substack{\text{expected epistemic uncertainty}\\ \text{after observation}}}
\label{eq:epistemic_value}
\end{equation}

Here, $H[\cdot]$ denotes entropy. Thus, $H[q(B_t)]$ represents uncertainty about the action--outcome mapping before observation, which we interpret as epistemic uncertainty about action--outcome relations. The term $\mathbb{E}_{q(s_{t+1}\mid u_t)}[H[q(B_t\mid s_{t+1},u_t)]]$ represents the epistemic uncertainty expected to remain after observing the state resulting from action $u_t$. Equation \eqref{eq:epistemic_value} therefore represents the difference between epistemic uncertainty before observation and the epistemic uncertainty expected to remain after observation. This formulation follows the definition of epistemic value in active inference.

\begin{figure}[!htbp]
\centering
\includegraphics[width=0.85\linewidth]{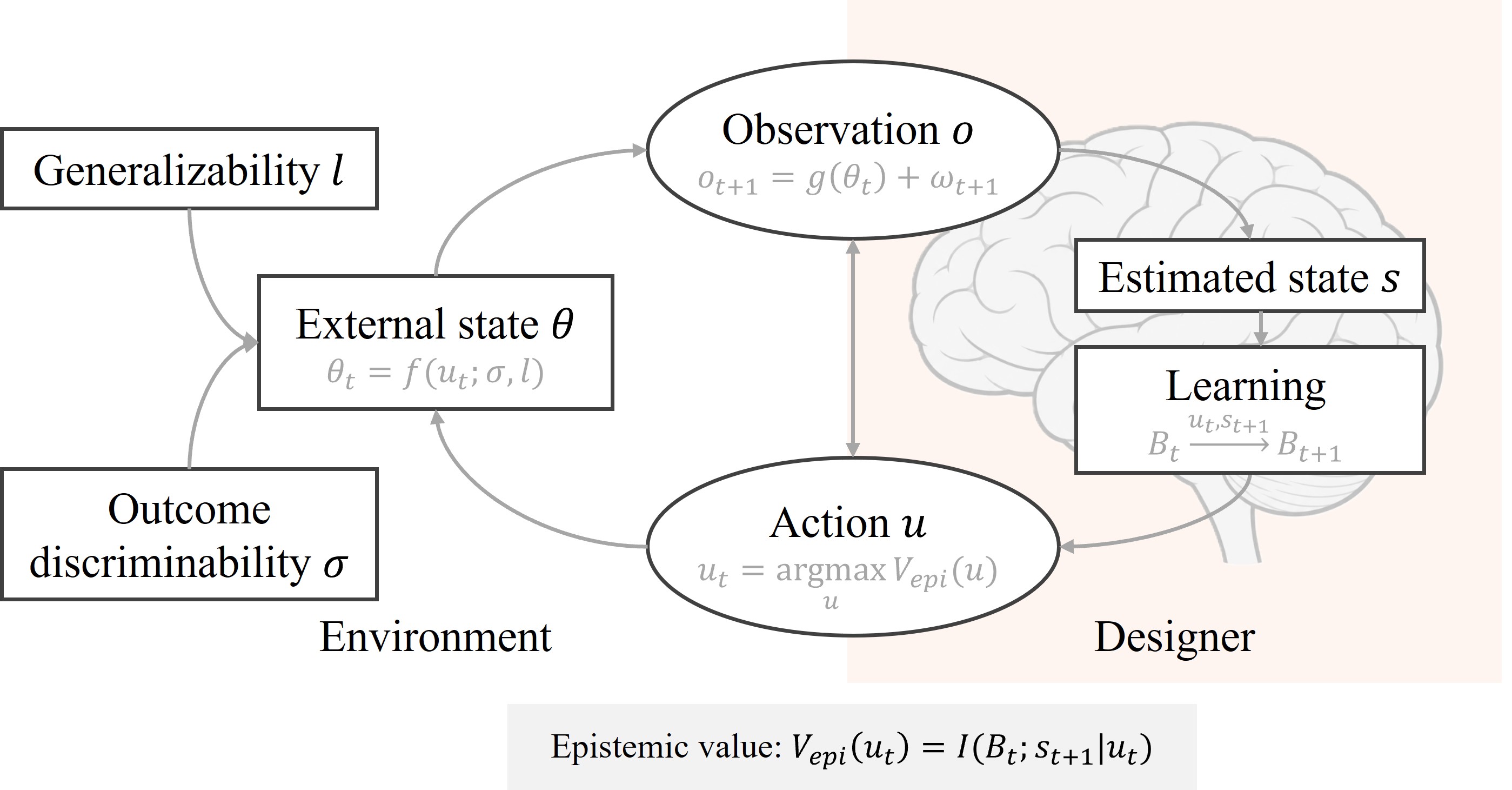}
\caption{
Conceptual diagram of the B-EUR model.
We designed this figure with reference to Figure 1 of \cite{Friston_2009}, which illustrates the perception--action loop in the free energy principle.
The environment is characterized by generalizability $l$ and outcome discriminability $\sigma$.
The designer accesses the external state $\theta_t$ through an action $u_t$ and obtains an observation $o_{t+1}$.
Based on the observation, the designer infers the internal state $s_{t+1}$ and updates the belief $B_t$ about action--outcome relations to $B_{t+1}$.
The designer then selects the next action based on epistemic value $V_{\mathrm{epi}}(u_t)=I(B_t;s_{t+1}\mid u_t)$.
The brain illustration on the right side of this figure was generated using ChatGPT (OpenAI).
}
\label{fig:b_eur_model}
\end{figure}

Figure \ref{fig:b_eur_model} shows an overview of the B-EUR model. We designed the figure with reference to the perception--action loop in the free energy principle, shown in Figure 1 of \cite{Friston_2009}. The environment is characterized by generalizability $l$ and outcome discriminability $\sigma$. The designer obtains an observation $o_t$ through an action $u_t$, updates the belief $B_t$ about action--outcome relations, and selects the next action based on the epistemic value $V_{\mathrm{epi}}$ of each action. Observation noise $\omega_{t+1}$ corresponds to aleatory uncertainty in design practice, such as fluctuations in user responses or accidental changes in the market environment, which learning cannot reduce.

When learning $B_t$ and estimating policies, we did not update only the observed location as a one-hot vector. Instead, we used Gaussian-weighted updates with a width proportional to the correlation length $l$, allowing each observation to update beliefs about neighboring locations. Studies on structured Multi-Armed Bandit tasks and function learning have shown that humans can understand input--output relations in an environment through limited exploration while generalizing observations to nearby options, even in large search spaces \citep{Wu_2018,Schulz_2020,Ke_2012}.

In the B-EUR model, epistemic uncertainty about the action--outcome mapping is formalized as the information-theoretic entropy of the matrix $B_t$. Expected uncertainty reduction is formalized as the reduction in this entropy before and after observing an output value.

\subsubsection{Simulation Conditions}

The previous section formalized expected uncertainty reduction as epistemic value in the Graph-Shape Guessing Game. In the simulations, we varied the environmental parameters, amplitude $\sigma$ and correlation length $l$, and computed their effects on prior entropy $H[q(B_t)]$ and epistemic value $V_{\mathrm{epi}}(u_t)$.

At each observation step, the agent computed epistemic value for all candidate input values and selected actions according to a softmax policy \citep{Smith_2022}, so that input values with higher epistemic value were more likely to be chosen. The agent then updated the belief distribution $B_t$ based on the obtained observation. The agent made $T=10$ sequential observations.

We repeated this procedure for each combination of amplitude $\sigma$ and correlation length $l$. Because graph shapes vary with random sampling even under the same $\sigma$ and $l$, we ran Monte Carlo repetitions with different random seeds. For each episode, we computed the time series of prior entropy and the epistemic value corresponding to the selected input values. Appendix Table \ref{tab:simulation_conditions} lists the main hyperparameters used in the implementation.

\subsection{Simulation Results}

\begin{figure}[!htbp]
\centering
\includegraphics[width=0.95\linewidth]{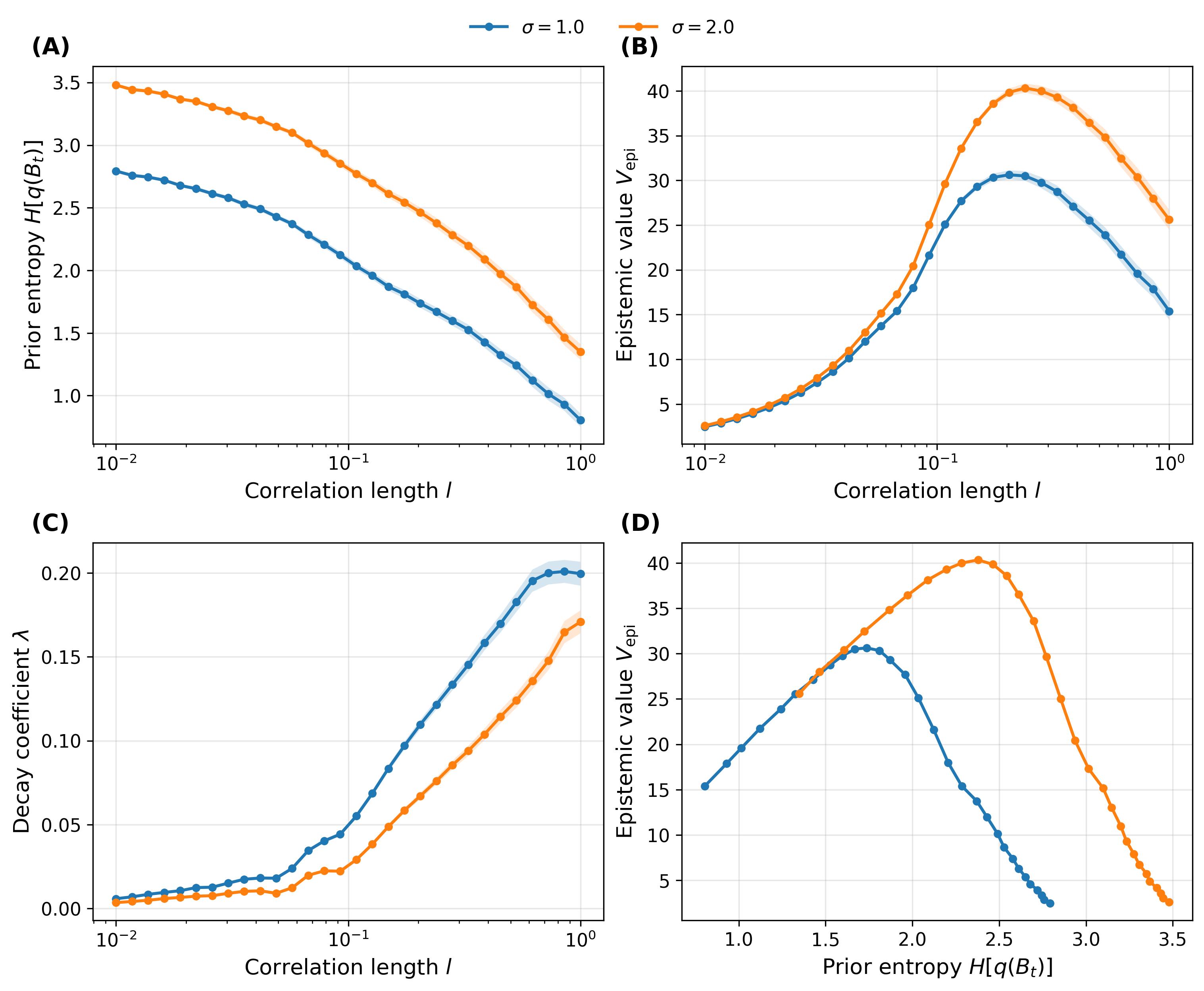}
\caption[Simulation results]{
Simulation results.
(A) Time-averaged prior entropy $H[q(B_t)]$. Prior entropy was higher for larger $\sigma$ and lower for larger $l$ (S1).
(B) Time-averaged epistemic value $V_{\mathrm{epi}}$. Epistemic value followed an inverted-U-shaped relationship with $l$ (S2) and increased with $\sigma$ (S3).
(C) Decay coefficient $\lambda$ obtained by fitting the exponential function $A\exp(-\lambda t)$ to the temporal change in epistemic value. The decay coefficient increased with $l$ and decreased with $\sigma$ (S4).
(D) Relationship between time-averaged prior entropy $H[q(B_t)]$ and time-averaged epistemic value $V_{\mathrm{epi}}$.
Shaded regions indicate standard errors based on Monte Carlo repetitions.
}
\label{fig:simulation_results}
\end{figure}

Figure \ref{fig:simulation_results} summarizes the simulation results. The simulations yielded four main predictions for the human experimental hypotheses.

\textbf{S1: Prior uncertainty increases with $\sigma$ and decreases with $l$.}
Prior entropy $H[q(B_t)]$ increased with amplitude $\sigma$ and decreased with correlation length $l$ (Figure \ref{fig:simulation_results}(A)). This result suggests that larger amplitude $\sigma$, which operationalized outcome discriminability in this task, also increased the range of possible output values and therefore raised prior uncertainty. In contrast, larger $l$ lowered prior uncertainty by increasing generalization across neighboring inputs.

\textbf{S2: Epistemic value follows an inverted-U-shaped relationship with $l$.}
Epistemic value $V_{\mathrm{epi}}$ showed an inverted-U-shaped relationship with $l$ and reached its maximum at an intermediate $l$ (Figure \ref{fig:simulation_results}(B)). This result suggests that epistemic value is low both when observations do not generalize to neighboring inputs and when a small number of observations is sufficient to learn the relation.

\textbf{S3: Epistemic value increases with $\sigma$.}
Epistemic value $V_{\mathrm{epi}}$ was higher under larger $\sigma$ conditions (Figure \ref{fig:simulation_results}(B)). This result suggests that trials provide more information when outcome differences among candidates are more distinguishable.

\textbf{S4: Epistemic value decays faster when $l$ is large and more slowly when $\sigma$ is large.}
We fitted an exponential function $A\exp(-\lambda t)$ to the temporal change in epistemic value $V_{\mathrm{epi}}$. The decay coefficient $\lambda$ increased with $l$ and decreased with $\sigma$ (Figure \ref{fig:simulation_results}(C)). This result suggests that epistemic value declines rapidly with observation in environments with high generalizability, whereas it declines more gradually under larger $\sigma$ conditions, where possible outcomes remain more widely separated.

Epistemic value $V_{\mathrm{epi}}$ also showed an inverted-U-shaped relationship with prior entropy $H[q(B_t)]$ (Figure \ref{fig:simulation_results}(D)). That is, epistemic value was low when uncertainty was too low or too high, and it was high at an intermediate level of prior uncertainty.

\begin{figure}[!htbp]
\centering
\includegraphics[width=0.60\linewidth]{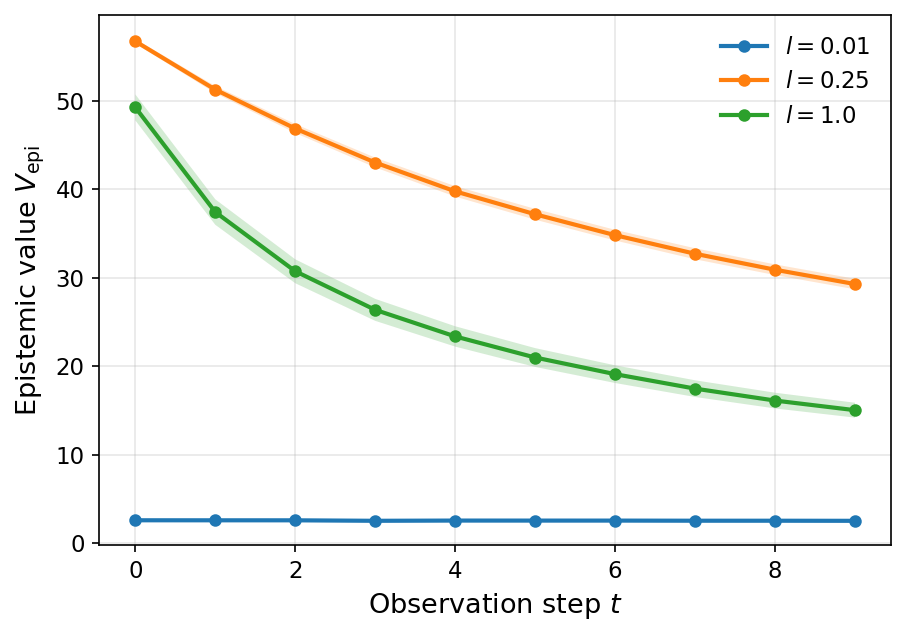}
\caption[Temporal change in epistemic value]{
Temporal change in epistemic value.
We fixed amplitude at $\sigma=2.0$ and set correlation length to $l=0.01, 0.25, 1.0$.
Epistemic value decreased with the number of observations, but it was already low at the initial step under $l=0.01$ and decreased rapidly under $l=1.0$.
In contrast, epistemic value remained relatively high under $l=0.25$.
Each line represents the mean across Monte Carlo repetitions, and shaded regions indicate standard errors.
}
\label{fig:mi_temporal}
\end{figure}

Figure \ref{fig:mi_temporal} shows the temporal change in epistemic value for representative conditions. When we fixed $\sigma=2.0$, $V_{\mathrm{epi}}$ decreased with the number of observations under all $l$ conditions. However, under $l=1.0$, it decreased rapidly in the early phase, whereas under $l=0.01$, it remained low from the initial step. In contrast, under $l=0.25$, epistemic value remained relatively high. This temporal pattern complements the inverted-U-shaped relationship between $l$ and epistemic value (S2) and the prediction that epistemic value declines faster when $l$ is large (S4).

Based on these results, the human experiments examined how perceived uncertainty, the subjective value of trying, and enjoyment changed with the generalizability and outcome discriminability of the environment.

\FloatBarrier

\section{Human Experiments}
\subsection{Method}

\subsubsection{Overview}

Experiment 1 examined how environmental properties affected subjective ratings of perceived uncertainty and the value of trying. Experiment 2 examined which environmental properties influenced participants' choices among exploration conditions.

We conducted the experiment at the Design Engineering Laboratory, The University of Tokyo, Japan. We used a PC web application implemented in React, and participants performed the Graph-Shape Guessing Game on a laboratory PC (Surface Laptop 4). We used a Logicool G gaming mouse G203-BK. The combined duration of Experiments 1 and 2 was approximately 70 minutes. Participants received a 2,000-yen online store gift card as compensation.

\subsubsection{Participants}

We recruited 46 adults aged between 20 and 39 years who had no visual diseases or impairments. We recruited participants through an experimental participant recruitment website (\url{jikken-baito.com}). All 46 participants completed the experiment.

We excluded two participants because of inconsistencies in the instruction procedure or problems in task performance. The final sample therefore included 44 participants (23 men and 21 women).

The Research Ethics Committee of the Graduate School of Engineering, The University of Tokyo, approved this study (approval number: KE26-13). Before the experiment, we explained the purpose of the study, experimental procedures, data to be collected, handling of personal information, and voluntary nature of participation. We obtained written informed consent from all participants. We also explained that participants could withdraw from the experiment at any time without disadvantage.

\subsubsection{Stimuli}

\begin{figure}[!htbp]
\centering
\includegraphics[width=0.95\linewidth]{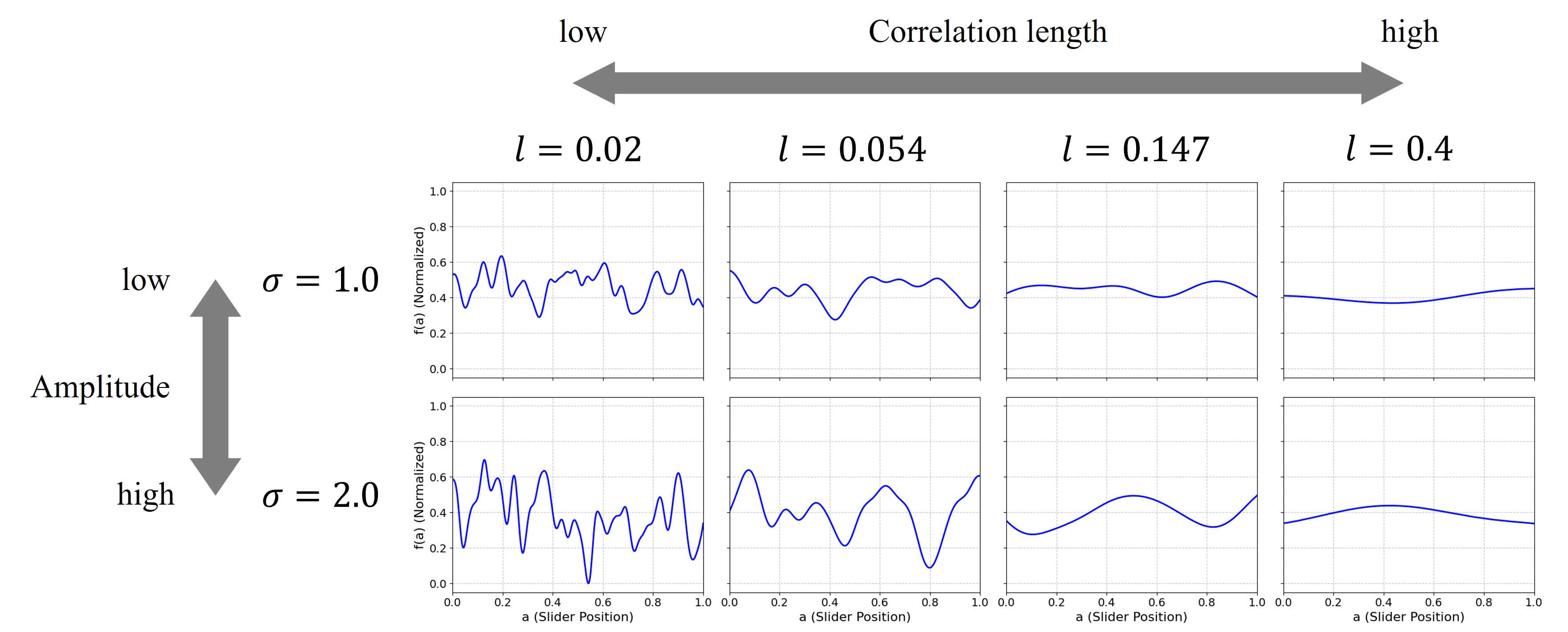}
\caption{
Examples of stimulus conditions used in the Graph-Shape Guessing Game.
The stimuli consisted of eight conditions defined by two levels of amplitude $\sigma$ and four levels of correlation length $l$.
A larger $\sigma$ increased the vertical variation of the graph and therefore increased outcome discriminability.
A smaller $l$ made the graph vary more rapidly and reduced generalizability, whereas a larger $l$ made the graph smoother and increased generalizability.
}
\label{fig:graph_stimuli}
\end{figure}

The stimuli consisted of graphs from eight conditions, defined by two levels of amplitude $\sigma$ and four levels of correlation length $l$ (Figure \ref{fig:graph_stimuli}). We set amplitude to $\sigma=1.0$ and $\sigma=2.0$. We divided the range from $0.02$ to $0.4$ into four levels on a logarithmic scale and set correlation length to $l=0.020, 0.054, 0.147, 0.400$. We used four levels of $l$ to test the predicted inverted-U-shaped relationship with epistemic value, and we used two levels of $\sigma$ to reduce participants' workload because we expected a monotonic effect of $\sigma$. As shown in Figure \ref{fig:graph_stimuli}, graphs with larger $\sigma$ had a wider vertical range and a broader range of possible output values. In contrast, graphs with smaller $l$ changed more rapidly along the input axis, so nearby input values tended to produce more different output values.

For stimulus generation, we fixed $\sigma$ and $l$ for each of the eight conditions and randomly generated individual graph shapes. Thus, even when graphs shared the same $\sigma$ and $l$, the graph shapes presented in practice trials, main trials, and choice screens did not overlap.

\subsubsection{Procedure}

\begin{figure}[!htbp]
\centering
\includegraphics[width=0.90\linewidth]{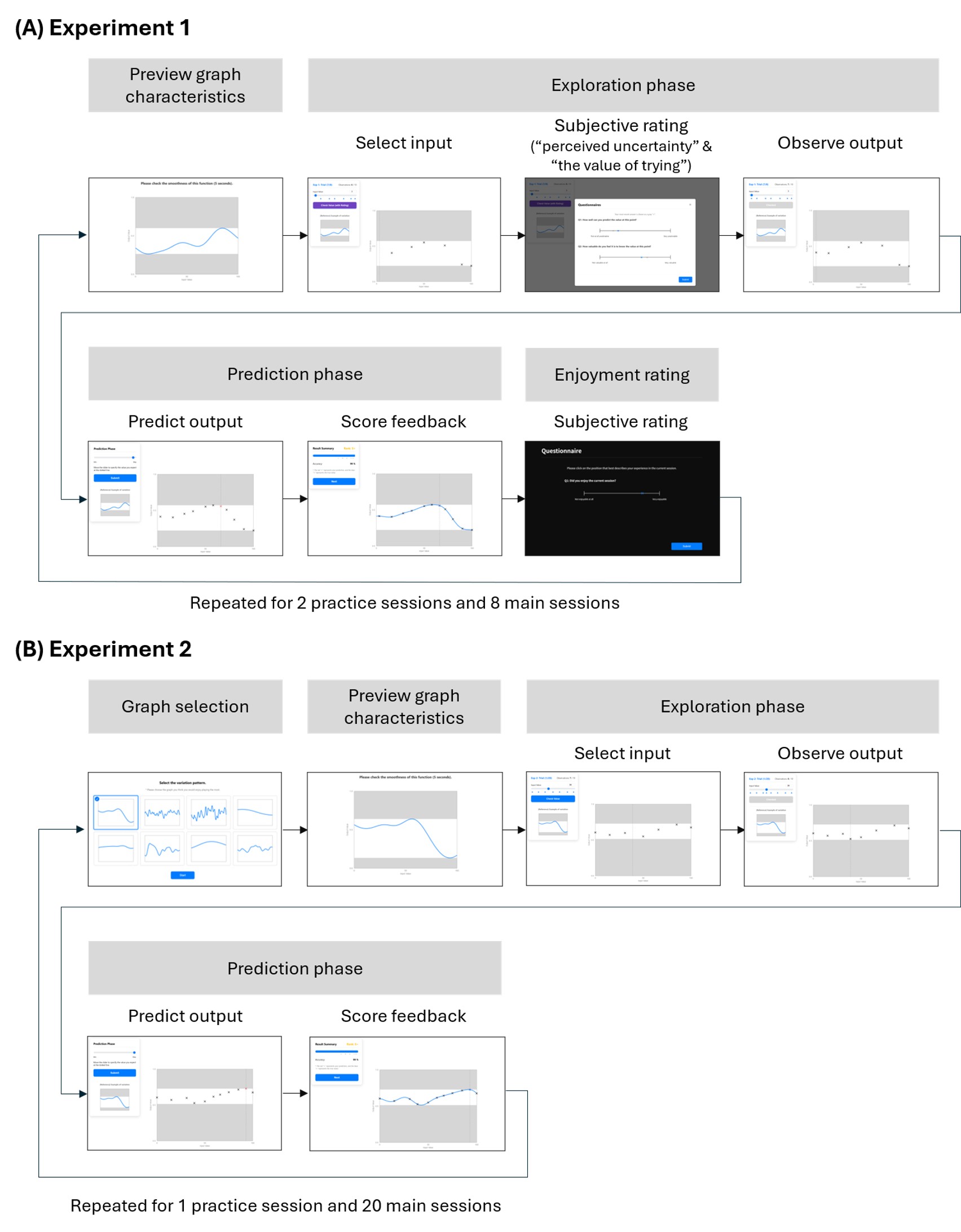}
\caption{
Experimental procedure and screen transitions within each session.
(A) Screen transitions in Experiment 1. Participants worked on each graph condition and provided subjective ratings during exploration and after each session.
(B) Screen transitions in Experiment 2. Participants selected the condition that looked most enjoyable from examples of the eight graph conditions and then explored a different graph with the same condition.
The experiment used Japanese screens, but this figure translates the on-screen text into English.
}
\label{fig:procedure_and_screen}
\end{figure}

Figure \ref{fig:procedure_and_screen} shows the experimental procedure and screen transitions within each session. The experiment consisted of Experiment 1, which collected subjective ratings during exploration, and Experiment 2, which measured participants' choices of graph conditions to explore. Throughout Experiments 1 and 2, we instructed participants that their goal was to enjoy the game.

\paragraph{Common session structure.}
Each session consisted of four phases: checking the graph features, exploration, prediction, and result confirmation. In the exploration phase, participants selected an input value using a slider and pressed the ``Confirm'' button to observe the output value at that location. Participants made 10 observations in each session. In the prediction phase, participants predicted the output value at a specified input value. The system then displayed a rank from S+ to D according to prediction accuracy. The rank did not change participants' reward.

\paragraph{Experiment 1: subjective ratings.}
In Experiment 1, participants completed one session for each of the eight graph conditions defined by $\sigma$ and $l$. Experiment 1 consisted of two practice sessions and eight main sessions. As shown in Figure \ref{fig:procedure_and_screen}(A), we collected subjective ratings during the exploration phase and after each session. At each observation step, participants answered questions about perceived uncertainty and the value of trying before pressing the ``Confirm'' button. After each session, participants answered a question about enjoyment. We measured enjoyment as an auxiliary indicator of the value of the exploration process itself, excluding external rewards. Because the task did not include performance-based rewards, we assumed that enjoyment partly reflected how worthwhile participants found the exploration process. Participants answered all questions using a Visual Analogue Scale (VAS).

\begin{itemize}
\item Perceived uncertainty: ``How well can you predict the value of this point?'' We measured this rating 10 times in each session. We reverse-coded the responses so that higher values indicated lower predictability, or higher perceived uncertainty.
\item Value of trying: ``How valuable do you feel it is to know the value of this point?'' We measured this rating 10 times in each session. We treated this rating as a subjective indicator corresponding to expected uncertainty reduction in the B-EUR model.
\item Enjoyment: ``Was this session enjoyable?'' We measured this rating once after each session.
\end{itemize}

\paragraph{Experiment 2: choice behavior.}
In Experiment 2, we measured which graph condition, defined by the combination of $\sigma$ and $l$, participants selected as the one that looked most enjoyable. Experiment 2 consisted of one practice session and 20 main sessions. As shown in Figure \ref{fig:procedure_and_screen}(B), each session began with a choice phase in which participants selected one graph condition from examples of the eight conditions. Participants selected the graph condition that they felt looked most enjoyable. They then explored a different graph that had the same $\sigma$ and $l$ as the graph shown in the choice and feature-check screens.

\subsubsection{Experimental Hypotheses}

We formulated the experimental hypotheses based on the simulation results of the B-EUR model. We assumed that perceived uncertainty corresponds to prior entropy $H[q(B_t)]$ in the model, and that the value of trying and choice behavior correspond to epistemic value $V_{\mathrm{epi}}(u_t)$. We treated enjoyment not as epistemic value itself, but as an auxiliary indicator of the subjective value of the exploration process without external rewards. Table \ref{tab:hypotheses} lists the hypotheses for the human experiments. Table \ref{tab:simulation_hypothesis_mapping} further summarizes how each hypothesis relates to the simulation predictions and their theoretical meanings.

\begin{table}[!htbp]
\centering
\caption{Hypotheses for the human experiments}
\label{tab:hypotheses}
\setlength{\tabcolsep}{3pt}
\begin{tabular}{P{0.10\linewidth}P{0.42\linewidth}P{0.22\linewidth}P{0.18\linewidth}}
\hline
Label & Hypothesis & Measure & Correspondence in the B-EUR model \\
\hline
H1-1 & A larger amplitude $\sigma$ increases the average perceived uncertainty.
& Subjective report of perceived uncertainty (Experiment 1)
& Prior entropy $H[q(B_t)]$ \\
H1-2 & A larger correlation length $l$ decreases the average perceived uncertainty.
& & \\
\hline
H2-1 & A larger amplitude $\sigma$ increases the average value of trying.
& Subjective report of the value of trying (Experiment 1)
& Epistemic value $V_{\mathrm{epi}}(u_t)$ \\
H2-2 & The average value of trying follows an inverted-U-shaped relationship with correlation length $l$.
& & \\
H2-3 & A larger amplitude $\sigma$ decreases the decay rate of the value of trying.
& & \\
H2-4 & A larger correlation length $l$ increases the decay rate of the value of trying.
& & \\
\hline
H3-1 & A larger amplitude $\sigma$ increases the average enjoyment.
& Subjective report of enjoyment (Experiment 1)
& Auxiliary indicator of the subjective value of the exploration process based on epistemic value \\
H3-2 & The average enjoyment follows an inverted-U-shaped relationship with correlation length $l$.
& & \\
\hline
H4-1 & A larger amplitude $\sigma$ increases the frequency of choosing a graph condition.
& Choice frequency for each combination of $\sigma$ and $l$ (Experiment 2)
& Epistemic value $V_{\mathrm{epi}}(u_t)$ \\
H4-2 & The frequency of choosing a graph condition follows an inverted-U-shaped relationship with correlation length $l$.
& & \\
\hline
\end{tabular}
\end{table}

\begin{table}[!htbp]
\centering
\caption{Correspondence between B-EUR model simulation predictions and experimental hypotheses}
\label{tab:simulation_hypothesis_mapping}
\setlength{\tabcolsep}{3pt}
\footnotesize
\begin{tabular}{P{0.16\linewidth}P{0.24\linewidth}P{0.15\linewidth}P{0.20\linewidth}P{0.15\linewidth}}
\hline
Simulation prediction & Interpretation & Mathematical index in the B-EUR model & Experimental indicator & Corresponding hypotheses \\
\hline
S1: Prior entropy increases with $\sigma$ and decreases with $l$
& Larger $\sigma$ broadens possible outputs and raises prior uncertainty, whereas higher generalizability lowers prior uncertainty
& $H[q(B_t)]$
& Perceived uncertainty
& H1-1, H1-2 \\

S2: Epistemic value follows an inverted-U-shaped relationship with $l$
& Trials are most informative when observations are neither too isolated nor too redundant
& $V_{\mathrm{epi}}(u_t)$
& Value of trying, enjoyment (auxiliary indicator), choice behavior
& H2-2, H3-2, H4-2 \\

S3: Epistemic value increases with $\sigma$
& More distinguishable outcome differences provide more information from a trial
& $V_{\mathrm{epi}}(u_t)$
& Value of trying, enjoyment (auxiliary indicator), choice behavior
& H2-1, H3-1, H4-1 \\

S4: Epistemic value decays faster with larger $l$ and more slowly with larger $\sigma$
& Higher generalizability reduces uncertainty with fewer observations, whereas higher outcome discriminability leaves more room for learning
& Temporal change in $V_{\mathrm{epi}}(u_t)$
& Decay rate of the value of trying
& H2-3, H2-4 \\
\hline
\end{tabular}
\end{table}

\subsubsection{Statistical Analysis}

We conducted the statistical analyses in R. For Experiment 1, we used linear mixed models (LMMs) for subjective ratings of perceived uncertainty, the value of trying, and enjoyment while accounting for participant-level differences. We standardized each subjective rating within each participant to correct for individual differences in scale use.

For perceived uncertainty, we examined the effects of amplitude $\sigma$, correlation length $l$, and number of observations. For the value of trying, we examined the effects of amplitude $\sigma$, correlation length $l$, the quadratic term of correlation length, and number of observations. To test the hypotheses about the decay rate of the value of trying, we also examined the interactions of number of observations with amplitude $\sigma$ and correlation length $l$. For enjoyment, which participants rated only once after each session, we examined the effects of amplitude $\sigma$, correlation length $l$, and the quadratic term of correlation length, without including number of observations.

For Experiment 2, we treated each trial as a discrete choice task in which participants selected one option from eight conditions. We applied a generalized linear mixed model (GLMM) to long-format data at the option level. The dependent variable indicated whether each option was selected, and the fixed effects included amplitude $\sigma$, correlation length $l$, and the quadratic term of correlation length. We mainly used the \texttt{lme4} package for model estimation. 
\subsection{Results}

\subsubsection{Experiment 1: Subjective Ratings}

\begin{figure}[!htbp]
\centering
\includegraphics[width=0.95\linewidth]{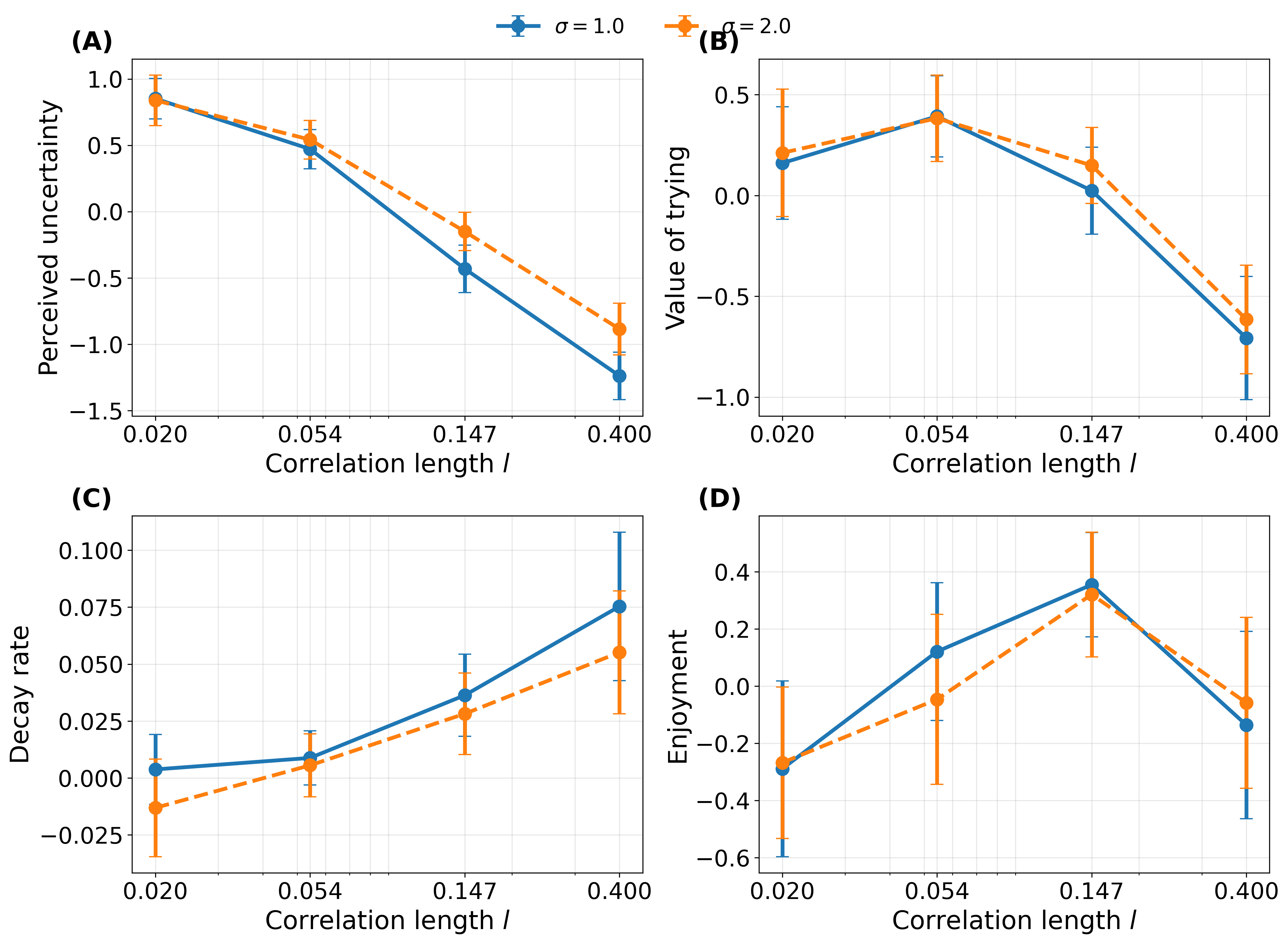}
\caption{
Results of subjective ratings in Experiment 1.
(A) Time-averaged perceived uncertainty. Perceived uncertainty was higher for larger $\sigma$ and lower for larger $l$.
(B) Time-averaged value of trying. The value of trying showed no clear effect of $\sigma$ and followed an inverted-U-shaped relationship with $l$.
(C) Decay rate of the value of trying. The value of trying decreased across observations, and the decrease was larger under larger $l$ conditions.
(D) Enjoyment. Enjoyment showed no clear effect of $\sigma$ and followed an inverted-U-shaped relationship with $l$.
The horizontal axis represents the levels of correlation length $l$, and colors represent the levels of amplitude $\sigma$.
}
\label{fig:exp1_ratings}
\end{figure}

Figure \ref{fig:exp1_ratings} shows the subjective ratings of perceived uncertainty, the value of trying, and enjoyment in Experiment 1.

To test H1-1 and H1-2, we applied an LMM with perceived uncertainty as the dependent variable and amplitude $\sigma$, correlation length $l$, and number of observations as fixed effects (Figure \ref{fig:exp1_ratings}(A)). The analysis showed a significant main effect of amplitude $\sigma$ ($\beta=0.106$, $SE=0.023$, $t=4.598$, $p<.001$), indicating that larger $\sigma$ increased perceived uncertainty. This result supports H1-1. The analysis also showed a significant main effect of correlation length $l$ ($\beta=-0.441$, $SE=0.010$, $t=-42.743$, $p<.001$), indicating that larger $l$ decreased perceived uncertainty. This result supports H1-2. The main effect of number of observations was also significant ($\beta=-0.167$, $SE=0.012$, $t=-14.361$, $p<.001$), indicating that perceived uncertainty decreased as participants made more observations.

To test H2-1 and H2-2, we analyzed how the mean value of trying changed with $\sigma$ and $l$ (Figure \ref{fig:exp1_ratings}(B)). The main effect of amplitude $\sigma$ was not significant ($\beta=-0.030$, $SE=0.054$, $t=-0.565$, $p=.572$), and H2-1 was not supported. In contrast, the linear term of correlation length $l$ was significant ($\beta=0.110$, $SE=0.024$, $t=4.579$, $p<.001$), and the quadratic term $l^2$ was negative and significant ($\beta=-0.161$, $SE=0.014$, $t=-11.196$, $p<.001$). Thus, the value of trying followed an inverted-U-shaped relationship with $l$, supporting H2-2.

To test H2-3 and H2-4, we analyzed changes in the value of trying across observations (Figure \ref{fig:exp1_ratings}(C)). The main effect of number of observations was significant ($\beta=-0.093$, $SE=0.018$, $t=-5.260$, $p<.001$), indicating that the value of trying decreased as participants made more observations. The $\sigma \times \mathrm{rep}_c$ interaction was not significant ($\beta=0.010$, $SE=0.010$, $t=0.988$, $p=.323$), and H2-3 was not supported. In contrast, the $l \times \mathrm{rep}_c$ interaction was significant ($\beta=-0.056$, $SE=0.004$, $t=-12.558$, $p<.001$), indicating that larger $l$ produced a larger decrease in the value of trying across observations. This result supports H2-4.

To test H3-1 and H3-2, we analyzed enjoyment as an auxiliary indicator of the subjective value of the exploration process (Figure \ref{fig:exp1_ratings}(D)). The main effect of amplitude $\sigma$ was not significant ($\beta=-0.026$, $SE=0.098$, $t=-0.261$, $p=.794$), and H3-1 was not supported. In contrast, the quadratic term $l^2$ was negative and significant ($\beta=-0.188$, $SE=0.049$, $t=-3.839$, $p<.001$), indicating that enjoyment followed an inverted-U-shaped relationship with $l$. This result supports H3-2. The linear term of $l$ showed a marginal trend ($\beta=0.084$, $SE=0.044$, $t=1.930$, $p=.054$).

Overall, in Experiment 1, perceived uncertainty increased with amplitude $\sigma$ and decreased with correlation length $l$. In contrast, the value of trying and enjoyment showed no significant main effect of $\sigma$, but both followed inverted-U-shaped relationships with $l$. The value of trying also decreased across observations, and this decrease was larger under larger $l$ conditions.

\subsubsection{Experiment 2: Choice Behavior}

\begin{figure}[!htbp]
\centering
\includegraphics[width=0.55\linewidth]{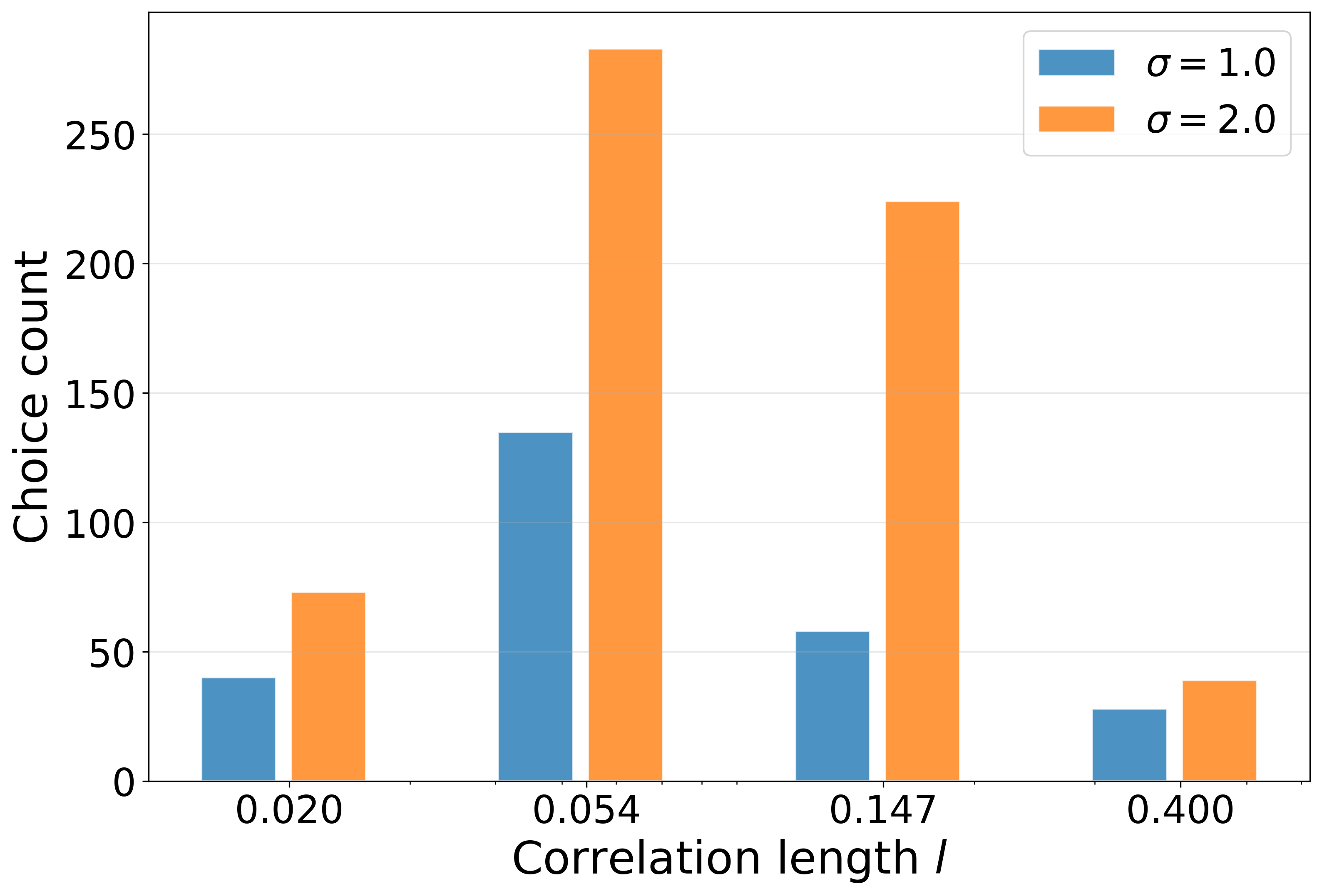}
\caption{
Choice frequency distribution for each condition in Experiment 2.
Participants more often selected conditions with larger $\sigma$ and intermediate $l$.
The horizontal axis represents the levels of correlation length $l$, and colors represent the levels of amplitude $\sigma$.
}
\label{fig:human_choice_dist}
\end{figure}

Figure \ref{fig:human_choice_dist} shows the distribution of choice frequencies for each combination of $\sigma$ and $l$. To test H4-1 and H4-2, we applied a GLMM with whether each condition was chosen as the dependent variable and amplitude $\sigma$, correlation length $l$, and $l^2$ as fixed effects.

The analysis showed a significant main effect of amplitude $\sigma$ ($\beta=1.035$, $SE=0.080$, $z=12.906$, $p<.001$), indicating that participants more often selected conditions with larger $\sigma$. This result supports H4-1. The quadratic term of correlation length $l$ was negative and significant ($\beta=-0.816$, $SE=0.046$, $z=-17.834$, $p<.001$), indicating that choice probability followed an inverted-U-shaped relationship with $l$. This result supports H4-2. The linear term of $l$ was also significant ($\beta=-0.275$, $SE=0.046$, $z=-5.934$, $p<.001$).

Overall, in Experiment 2, choice probability was higher for larger $\sigma$ conditions and followed an inverted-U-shaped relationship with correlation length $l$.

\FloatBarrier

\section{Discussion}

\subsection{Summary of findings}

In this study, we focused on one aspect of design exploration: learning about action--outcome relations through trials. Within this scope, we formalized the value of trying a design option as expected uncertainty reduction. The B-EUR model predicted that prior epistemic uncertainty would increase with amplitude $\sigma$ and decrease with correlation length $l$. It also predicted that epistemic value would follow an inverted-U-shaped relationship with $l$ and increase with $\sigma$.

The human experiments showed that correlation length $l$, which operationalized generalizability, was consistently related to indicators of the value of trying. The value of trying and enjoyment followed inverted-U-shaped relationships with $l$, and participants more often selected conditions with intermediate $l$. Amplitude $\sigma$, which operationalized outcome discriminability, affected perceived uncertainty and choice behavior, whereas the subjective ratings of the value of trying and enjoyment did not show clear effects of $\sigma$. We discuss this difference after presenting the implications for design practice.

These findings provide a computational account of one mechanism relevant to the UDA model's open question concerning action selection. B-EUR evaluates candidate actions according to their expected reduction of uncertainty about action--outcome relations. Within this scope, the model does not describe the full progression or combination of information, knowledge-sharing, and representation actions proposed in the UDA model.

\subsection{What environmental properties make design options worth trying?}

The B-EUR model and the human experiments suggest that moderate generalizability consistently increases the value of trying design actions or candidate solutions. Outcome discriminability was reflected more clearly in perceived uncertainty and relative choice behavior than in subjective ratings during exploration.

\textbf{P1: Moderate generalizability}
Trial outcomes need to generalize to other candidates to some extent. If generalizability is too low, designers cannot use the obtained information to predict neighboring candidates. If generalizability is too high, different candidates provide highly overlapping information, and additional exploration contributes little information.

\textbf{P2: Sufficient outcome discriminability}
Outcome discriminability refers to how clearly different design options produce distinguishable outcomes after a design activity, such as prototyping, evaluation, or review. Clearer outcome differences can make differences among exploration conditions more salient and may help designers infer which design differences led to which outcomes.

We therefore interpret moderate generalizability as a robust condition for increasing the value of trying. Sufficient outcome discriminability may further support exploration, particularly when designers compare candidate conditions relative to one another. The inverted-U-shaped relationships of the value of trying and enjoyment with generalizability are broadly consistent with arousal potential theory, which proposes that an intermediate level of arousal induced by novelty, complexity, and uncertainty maximizes positive hedonic responses \citep{Berlyne_1970,Yanagisawa_2021,Yanagisawa_2025}. The observed inverted-U-shaped pattern also aligns with curiosity research suggesting that exploration depends on learnability and the possibility of progress in understanding \citep{Ten_2020,Poli_2022,Singh_2021,Buyalskaya_2020}. We next extend these insights to prototyping, problem framing, and organizational support in design practice.

\subsection{Implications for Design Practice: Driving Designers' Exploration}

We connect P1, moderate generalizability, and P2, sufficient outcome discriminability, to design practice from three perspectives: prototyping, problem framing, and design management.

\subsubsection{Implications for prototyping: designing informative comparisons}

Prototyping is an iterative process in which designers materialize early ideas, obtain feedback from users and stakeholders, and reflect the results in subsequent design decisions \citep{Elkoutbi}. From the perspective of this study, prototyping does not only evaluate whether individual options are good or bad. It also helps designers learn the relation between design options and responses.

From the perspective of P1, designers should focus on design elements that create differences in evaluation or response while allowing those differences to generalize to similar options. Here, $l$ represents how far the relation between prototype differences and user evaluations or responses generalizes, depending on the design element under consideration. If designers change an element with too large an $l$, the obtained information can become redundant. If they change an element with too small an $l$, even small differences can produce large changes in response, making it difficult to find a learnable relation. Thus, to obtain prototyping outcomes that are informative for subsequent design decisions, designers should compare prototype sets that differ in design elements with moderate generalizability. For example, when examining a new automobile interior, changing only details that have little influence on user responses, such as screw hole positions, has low learning value. Conversely, changing many elements at once, such as dashboard shape, material, color scheme, and control layout, makes it difficult to learn which difference produced the response. Focusing on one element, such as control layout, that produces response differences while generalizing to similar options can help designers obtain useful knowledge for the next design decision.

From the perspective of P2, the way designers collect user feedback also matters. Binary evaluations such as ``good'' or ``bad'' may fail to capture differences in responses among prototypes. In contrast, multiple evaluation dimensions, such as ``innovative,'' ``reassuring,'' ``familiar,'' and ``luxurious,'' can help designers learn which differences lead to which responses. Feedback formats that increase outcome discriminability may make the relation between design options and responses easier to distinguish and provide more useful information for subsequent exploration.

\subsubsection{Implications for problem framing: bounding the design space for resolvable uncertainty}

Studies on problem--solution coevolution \citep{Dorst_2001,Parsons_2026,Cash_2023} and problem framing in design teams \citep{Litster_2024} show that design activity involves not only solving a given problem but also defining and reframing what the problem is. Research on creative problem solving also treats problem finding as a process as important as solution generation \citep{Wakefield_1985}. From the perspective of this study, we can understand problem framing as the act of carving out a search space in which designers can increase the value of trying.

From the perspective of P1, a problem scope that is too narrow or too broad lowers the epistemic value of exploration. If the problem scope is too narrow, designers can understand the relation after only a few trials and reduce uncertainty too early. If the problem scope is too broad, designers cannot easily generalize knowledge from one trial to other candidates, and exploration does not sufficiently improve understanding. For example, in a car model change, framing the problem as ``designing the future mobility experience'' may be too broad, whereas framing it as ``changing only the headlight shape of an existing model'' may be too narrow. Designers should limit the target users, value dimensions, and exploration targets to some extent while leaving room for comparisons among multiple options to update their understanding of the problem.

\subsubsection{Implications for managing exploratory design work}

Our findings also provide implications for organizational management that supports designers' exploration. Designers' decision autonomy can strengthen the relation between exploration and innovativeness, but it does not always translate directly into market performance \citep{Tabeau_2016}. Creativity research suggests that constraints can support creative activity when they come with appropriate support \citep{Rosso_2014}. Critiques and reviews can also function not only as opportunities to accept or reject ideas but also as opportunities to indicate what designers should compare next to improve specific evaluations \citep{Christensen_2016}.

From this perspective, managers and project leaders should not simply give designers freedom. They should create environments in which exploration with high epistemic value can occur. They can support problem framing that makes the search space neither too narrow nor too broad, help designers prepare prototype sets with both commonalities and differences, and design evaluation dimensions and feedback formats.

For example, in a review of a new automobile interior design, if a leader simply judges an option as ``good'' or ``bad,'' the designer may struggle to decide what to explore next. In contrast, a more specific critique, such as ``this option feels reassuring but does not look innovative enough; next, keep the material fixed and compare options that vary only in control layout,'' can support the designer's next exploration. Such critique increases outcome discriminability (P2) by specifying multiple evaluation dimensions and also supports moderate generalizability (P1) by directing exploration toward a design element, such as control layout, with moderate generalizability.

\subsection{Why outcome discriminability was reflected differently in ratings and choices}

In Experiment 1, amplitude $\sigma$ increased perceived uncertainty but did not show significant effects on the value of trying or enjoyment. In contrast, in Experiment 2, participants more often selected conditions with larger $\sigma$. This difference may have occurred because, in this task, the value of trying depended more strongly on how well observations generalized to other inputs than on the output range itself. Correlation length $l$ mainly determined this generalizability. For example, if the hidden graph were a sine wave, changing the amplitude would not prevent participants from predicting unknown values once they understood the wavelength and phase. Similarly, in this task, $\sigma$ changed the output gain, but $l$ mainly determined how much participants could learn the input--output relation. Thus, the effects of $l$ may have dominated ratings of the value of trying and enjoyment, making the effect of $\sigma$ difficult to detect.

At the same time, $\sigma$ affected perceived uncertainty and choice behavior. For perceived uncertainty, larger $\sigma$ expanded the range of possible output values, especially in the early phase of exploration. When participants had observed only a few points and had not yet grasped the overall graph trend, larger $\sigma$ likely broadened the possible output values and increased subjective uncertainty.

Experiment 2 also presented examples of all eight graph conditions simultaneously and asked participants to choose the one that looked most enjoyable. Thus, compared with Experiment 1, in which participants evaluated one graph at a time, Experiment 2 may have made differences in amplitude, and thus in the distinguishability of outcome differences, more salient. In other words, $\sigma$ did not clearly affect absolute subjective ratings of the value of trying or enjoyment, but it may have influenced choice behavior when participants compared multiple conditions and perceived differences among possible outcomes more directly. 

Overall, our results suggest that $l$ is the main factor related to generalizability and the value of trying, whereas $\sigma$ broadens the range of possible output values and appears in perceived uncertainty and relative choice behavior.

\subsection{Limitations and future directions}

This study has several limitations. First, the Graph-Shape Guessing Game does not reproduce the full design process. It extracts one aspect of design exploration: learning how design actions produce outcomes that are informative for subsequent design decisions. Future work should extend this approach to tasks closer to design practice, such as prototyping tasks \citep{Camburn_2015,Hansen_2020}, user research \citep{Frich_2021}, and creativity tasks \citep{Erwin_2022}. Relatedly, the B-EUR model does not distinguish among or model transitions across the information, knowledge-sharing, and representation actions proposed in the UDA model. It abstracts a common structure in which a candidate action produces an observable outcome that updates beliefs about action--outcome relations.

Second, this study abstracted the relation between input and output as a one-dimensional function. In actual design, people evaluate products through multiple modalities, such as vision, touch, sound, and the feel of operation \citep{Spence_2011,Maki_2019,Miyazaki_2016}. Evaluation also involves multiple dimensions, such as functionality, usability, aesthetics, quality, and satisfaction \citep{Tajdini_2021,Brucks_2000,Christensen_2016,Mano_1993}. Although these modalities and dimensions could in principle be represented as multidimensional outcomes, the present model does not explicitly capture correlations or interactions among them. The meaning of the same observation can also change depending on the product category and use context \citep{Hu_2010,Berni_2024,De_Maeyer_2011}. Future work should extend the model to include multidimensional and multimodal observations, their correlation structure, and context-dependent interpretations.

Third, actual design exploration depends not only on epistemic value but also on pragmatic value, resource constraints, expertise, and organizational culture. This study did not introduce preferences over output values, and it examined only the effect of epistemic value on exploration. In practice, however, designers pursue both exploration to reduce uncertainty and exploitation to obtain desired outcomes \citep{March_1991,Friston_2017}. Exploration can also change depending on resource constraints \citep{Letting_2025,Savage_1998,Chaudhari_2020}, expertise and domain knowledge \citep{Tseng_2011}, and organizational cultures that support or suppress exploration \citep{Elsbach_2018}. Future work should model these factors and test the framework across different scenarios, including designers with varying levels of expertise and different types of projects, such as market-oriented and research-oriented projects.

\section{Conclusion}

This study aimed to provide a computational account of what makes design options worth trying under uncertainty. Focusing on one aspect of design exploration---learning action--outcome relations through trials---we proposed the Bayesian Expected Uncertainty Reduction (B-EUR) model, which formalizes the value of a candidate action as its expected reduction of epistemic uncertainty. Simulations and human experiments examined two environmental properties: generalizability and outcome discriminability. The results consistently showed that the value of trying was highest when trial outcomes generalized moderately to neighboring candidates. Outcome discriminability increased epistemic value in the simulations and was reflected in perceived uncertainty and relative choice behavior, although it did not clearly affect subjective ratings of the value of trying or enjoyment during exploration.

The B-EUR model offers three main contributions to design research. First, it formalizes the value of trying a design option as expected uncertainty reduction, providing a computational mechanism for evaluating candidate actions before selection. In this respect, the model addresses one part of the UDA model's open question concerning how changes in uncertainty perception determine action selection, while not attempting to model the full progression or combination of information, knowledge-sharing, and representation actions. Second, the study identifies environmental conditions that shape the informational value of design exploration: trial outcomes should generalize sufficiently to inform neighboring candidates without becoming redundant, and outcome differences should be sufficiently distinguishable to support comparison and learning. Third, these findings provide a basis for designing more informative exploration conditions through the construction of prototype sets, the framing of design problems, and the organization of evaluation and feedback. Together, the B-EUR model and the empirical findings connect uncertainty-driven accounts of design activity with a computational explanation of how candidate design actions can be valued in terms of their expected learning outcomes.

\section*{Financial support}

This work was supported by The University of Tokyo ``Advanced AI Talent Development to Lead the Next-Generation AI for Intelligent Society (BOOST NAIS)'', supported by the Broadening Opportunities for Outstanding young researchers and doctoral students in STrategic areas (BOOST) of the Japan Science and Technology Agency (JST).

\section*{Supplementary materials}

The anonymized experimental data, B-EUR model simulation code, and statistical analysis scripts supporting the findings of this study are available in an \href{https://osf.io/kgmh5/overview?view_only=d0b069b4cd1f4db780e43edce7628f52}{\textcolor{blue}{\underline{online repository}}}.

\section*{Declaration of generative AI and AI-assisted technologies in the manuscript preparation process}

During the preparation of this work, the authors used OpenAI ChatGPT to support translation, language editing, and the creation of illustrative elements in figures. Specifically, ChatGPT was used to generate the central human illustration in Figure \ref{fig:graph_shape_game} and the brain illustration on the right side of Figure \ref{fig:b_eur_model}. The authors integrated these illustrative elements into the figures, reviewed and edited all AI-assisted content as needed, and take full responsibility for the content of the manuscript.

\bibliographystyle{plainnat}
\bibliography{references}

\clearpage
\appendix
\section*{Appendices}

\subsection*{Appendix A: Simulation Parameters}
\label{app:simulation-parameters}

Table \ref{tab:simulation_conditions} lists the main constants and hyperparameters used in the simulations.
In the Gaussian-weighted belief update, we used an internal update width $l_{\mathrm{model}}$ to determine how far each observation updated neighboring inputs.
This width was set proportional to the environmental correlation length $l$ as $l_{\mathrm{model}}=\beta l$.

\begin{table}[!htbp]
\centering
\caption{Main constants and hyperparameters used in the simulations}
\label{tab:simulation_conditions}
\setlength{\tabcolsep}{4pt}
\begin{tabular}{P{0.22\linewidth}P{0.18\linewidth}P{0.48\linewidth}}
\hline
Symbol/variable & Value & Description \\
\hline
$T$ & $10$ & Number of observations per episode \\
$N_{\mathrm{grid}}$ & $101$ & Number of candidate inputs in the input space $[0,1]$ \\
$\alpha_0$ & $0.5$ & Initial parameter of the Dirichlet distribution \\
$\Delta$ & $0.3$ & Discretization width of the output value $f$ \\
$n_{\mathrm{MC}}$ & $100$ & Number of Monte Carlo repetitions per condition \\
$\gamma$ & $1.0$ & Inverse temperature parameter of the softmax policy \\
$\eta$ & $1.0$ & Scale of belief updating by observation \\
$\beta$ & $3.0$ & Conversion coefficient from the environmental correlation length $l$ to the update width $l_{\mathrm{model}}$ \\
\texttt{use\_gaussian} & \texttt{True} & Setting for generalizing observations to neighboring inputs with Gaussian weights \\
\hline
\end{tabular}
\end{table}

\end{document}